\documentclass[12pt]{article}

\usepackage{arxiv}

\usepackage[utf8]{inputenc} % allow utf-8 input
\usepackage[T1]{fontenc}    % use 8-bit T1 fonts
\usepackage{hyperref}       % hyperlinks
\usepackage{url}            % simple URL typesetting
\usepackage{booktabs}       % professional-quality tables
\usepackage{amsfonts}       % blackboard math symbols
\usepackage{nicefrac}       % compact symbols for 1/2, etc.
\usepackage{microtype}      % microtypography
\usepackage{graphicx}
\usepackage{natbib}
\usepackage{doi}

\title{Estimating Remaining Lifespan from the Face}

\author{ \href{https://orcid.org/0000-0001-9749-3507}{\includegraphics[scale=0.06]{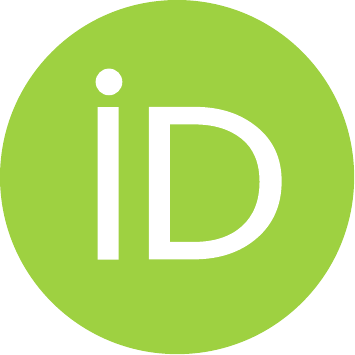}\hspace{1mm}Amir Fekrazad}\thanks{} \\
	College of Business\\
	Texas A\&M University-San Antonio\\
	San Antonio, TX 78224 \\
	\texttt{afekrazad@tamusa.edu} \\
}

\renewcommand{\shorttitle}{Estimating Remaining Lifespan from the Face}

\hypersetup{
pdftitle={Estimating Remaining Lifespan from the Face},
pdfsubject={},
pdfauthor={Amir Fekrazad},
pdfkeywords={Remaining Lifespan, Life expectancy, Convolutional Neural Networks, Computer Vision},
}

\begin{document}
\maketitle

\begin{abstract}
The face is a rich source of information that can be utilized to infer a person's biological age, sex, phenotype, genetic defects, and health status. All of these factors are relevant for predicting an individual's remaining lifespan. In this study, we collected a dataset of over 24,000 images (from Wikidata/Wikipedia) of individuals who died of natural causes, along with the number of years between when the image was taken and when the person passed away. We made this dataset publicly available. We fine-tuned multiple Convolutional Neural Network (CNN) models on this data, at best achieving a mean absolute error of 8.3 years in the validation data using VGGFace. However, the model's performance diminishes when the person was younger at the time of the image. To demonstrate the potential applications of our remaining lifespan model, we present examples of using it to estimate the average loss of life (in years) due to the COVID-19 pandemic and to predict the increase in life expectancy that might result from a health intervention such as weight loss. Additionally, we discuss the ethical considerations associated with such models.

\end{abstract}

% keywords can be removed

\section{Introduction}
Is it possible for an AI model to receive as input the facial image of a person and predict, with some degree of accuracy, how many years of life that person has left? Predicting remaining lifespan (RL) is a complex task that requires taking into account a wide range of factors, many of which can be inferred from the face. For instance, a person's biological age, as determined by genetics, health, and lifestyle, is often reflected in their appearance. Additionally, certain genetic defects, diseases, and health issues may be visible on the face. There are also differences in life expectancy among different sexes and racial/ethnic groups, which can potentially be determined by facial features.

If a human expert is asked to estimate the RL of a person solely from a photograph of their face, he or she will likely follow the following steps: first, guess the person's age from the facial features, second, find the average life expectancy of the sex and race cluster to which the person belongs, and finally, calculate the RL as the difference between the life expectancy and current age. Similarly, a deep learning model can perform this process within its "blackbox." These models have been shown to estimate age with a high level of accuracy \citep{AgboAjala2021}. Furthermore, when trained on a large and diverse dataset, such models can learn to cluster individuals not only by sex and race, but also by more subtle and fine-grained facial features and determine the average life expectancy for each cluster. Therefore, in theory, a deep learning model has the potential to exceed human expert performance in predicting RL from the face.

In this study, our goal was to develop an AI model that could use facial features to predict RL. To create a dataset for this task, we used Wikidata and Wikipedia to collect images of individuals who died of natural causes between 1990 and 2022 (inclusive), along with captions indicating when the images were taken. By comparing the date of death with the date the image was taken, we were able to determine the number of years each person lived after the image was taken, which we used as the label for each image. The images were then subjected to face detection and alignment before being used for training. The final dataset consisted of more than 24,000 images which were used to fine-tune deep convolutional neural network (CNN) models. In order to improve the accuracy of the model, we also applied data augmentation techniques such as flipping and cropping of images.

We experimented with various CNN models including FaceNet \citep{schroff2015facenet}, VGGFace \citep{parkhi2015deep}, and VGGFace2 \citep{cao2018vggface2}, all pretrained on large datasets of faces for the task of face recognition. These models map the face into a high-dimensional numerical vector (face embedding). We then fed these embeddings to a few fully-connected layers, culminating in a final layer that generates the output (RL). To generate the output, we considered classification (argmax of a softmax layer), regression (a fully-connected layer with 1 unit), and expected value (sum of values multiplied by their respective softmax probabilities), à la \citet{rothe2015dex}. The VGGFace model with two layers of 1024 fully-connected (FC) nodes followed by a regression output yielded the best performance, with a Mean Absolute Error (MAE) of 8.3 years.

While an MAE of 8.3 may not seem impressive at first glance, it is important to note that RL is inherently highly uncertain. In comparison, for age, which involves less uncertainty, state-of-the-art models have MAEs of approximately 2-3 years \citep{AgboAjala2021}. Our error analysis revealed that the younger the person is at the image, and hence the longer the expected RL, the less accurate the predictions of the model become. This is both because predicting the RL of a younger person is more difficult in nature and because our data includes relatively fewer examples of images taken at a very young age.

There are a variety of potential applications for RL prediction from the face. For example, life insurance companies can use such models alongside the conventional methods (mortality tables and physical evaluations) to estimate an individual's lifespan and determine insurance premiums accordingly. Additionally, RL prediction models can be used to estimate the loss of life, in terms of years (as opposed to just the number of people who died), resulting from deadly events. For example, using our model, we show that out of the 278 people in our dataset who died of COVID-19 (these examples were not used for training and validation), the average loss of life was 3.3 years, i.e. on average, they would have lived another 3.3 years had the pandemic not occurred. Furthermore, such models can be used to demonstrate how health interventions and lifestyle changes, such as weight loss, can impact an individual's RL. By applying the model to the pictures of celebrities before and after a significant weight loss, we can see an increase in RL. However, it is important to note that these predictions must be used with caution and in conjunction with other forms of analysis. 

The remainder of this paper is organized as follows: in Section \ref{sec:data}, we describe the process of collecting and cleaning the data. In Section \ref{sec:method}, we present the process of developing and training the model. Section \ref{sec:error-analysis} performs error analyses on the validations set predictions. Section \ref{fig:apps} demonstrates a few examples of how and where the model can be applied. Section \ref{sec:ethics} discusses the ethical implications of RL prediction models. Section \ref{sec:conclusion} provides concluding remarks.

\section{Data}
\label{sec:data}

To train a model capable of predicting remaining lifespan (RL) from a facial image, we need a dataset of images of individuals and the number of years the person lived after the image was taken. To create this dataset, we queried Wikidata and scraped Wikipedia, as outlined in the remainder of this chapter. We have made our dataset openly accessible for other researchers to utilize in their own studies and experimentation.\footnote{Link to download the remaining lifespan dataset (images and labels): \href{https://github.com/fekrazad/remaining-lifespan-ai}{github.com/fekrazad/remaining-lifespan-ai}}

An alternative suggestion might be to use profiles of deceased individuals on social media websites like Facebook to collect such data. However, it is important to keep in mind that such websites have only been around for the past 20 years and as a result, the RL variable collected from them would be truncated. Additionally, the data in social media websites may not be as reliable or accurate as data from other sources like Wikidata and Wikipedia.

We first queried Wikidata, a searchable knowledge graph, to find all persons who have both a date of birth and a date of death specified at least to the year level, died between 1990 and 2022 (inclusive), and whose manner of death is either "natural causes" or not specified. It is worth noting that while we do not require the date of birth in order to create the RL labels, it is useful for analyzing the sample and conducting error analysis.

In some cases, multiple dates of birth or death may be recorded for an individual due to contradictory sources or data entry errors. If the year of the conflicting dates differs, we excluded these entries from our dataset.

In Wikidata, there are two properties related to death: "manner of death", which is more general (e.g., natural causes, accident), and "cause of death", which is more specific (e.g., a particular disease, airplane crash). We only considered entries for which the manner of death is either "natural causes" or not specified. We then created a separate dataset that maps each cause of death to the most relevant manner of death. If a person's manner of death is not specified but the cause of death is specified and related to an unnatural cause, that entry was excluded from the dataset.

While dying from COVID-19 is technically considered death by a natural cause, since a global pandemic of this proportion is a once-in-a-century phenomenon, we excluded individuals whose cause of death was attributed to the virus.

We were able to easily obtain images with associated "point in time" properties from Wikidata entries. For entries without this property, we attempted to extract the year the image was taken by searching the caption for a single number in the 19** or 20** format. We assumed this number to be the year the image was taken.

Unfortunately, there is not a one-to-one correspondence between Wikidata entries and Wikipedia articles. To obtain images and their captions for Wikidata entries with an empty image field, we scraped the corresponding English Wikipedia page. If the page included an image and the image's caption included a single number in the 19** or 20** format, we assumed this to be the year the image was taken. If the image year falls before the individual's date of birth or after their date of death, we discarded the image as it is evidently not the correct year the image was taken.

\begin{figure}
  \centering
  \includegraphics[width=0.5\linewidth]{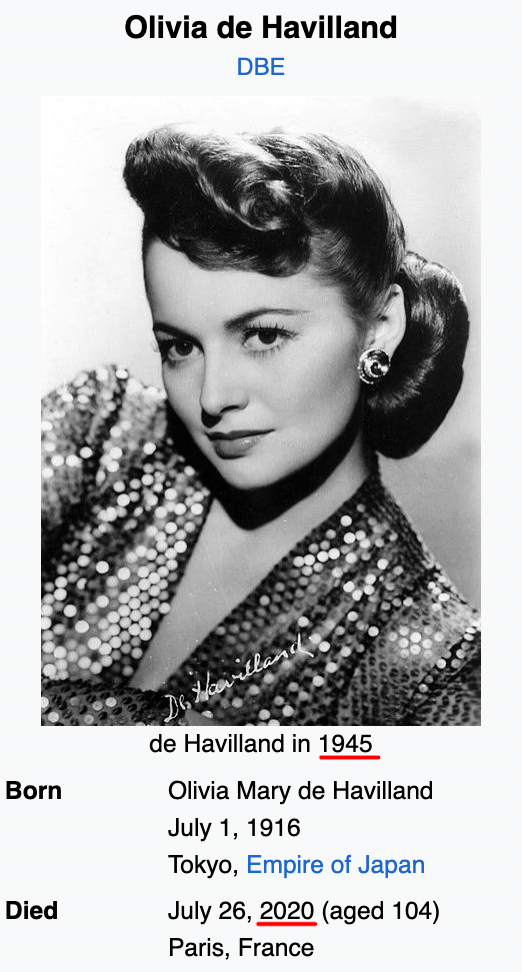}
  \caption{An example of successful data extraction from Wikidata/Wikipedia. The difference between the year the image was taken (1945) and the year of death (2020) gives us an RL label of 75 for the image.}
  \label{fig:wiki-example}
\end{figure}

\figurename~\ref{fig:wiki-example} shows an example of a Wikipedia page where the image and image year can be scraped. Combined with the year of death, we can determine the RL label for the image.

There are multiple potential sources of error in the data. First, since the death date and image date are measured in years, the RL variable for an image can be off by a year. For example, if the image was taken in January (December) 2000 and the person died in December (January) 2001, the actual RL is almost two (zero) while in our data it will be recorded as one. 

Additionally, since Wikidata/Wikipedia entries are created and edited by volunteers, they can be erroneous or incomplete. We assumed that if a person does not have a manner and cause of death specified, they died of natural causes. Upon manual inspection of a sample of entries, this turned out not to hold for those who died at younger ages. For this reason, we limited the data to those who were 50 years or older at the time of their death. Less than 5\% of the data fell under this threshold.

\begin{figure}
\centering
\includegraphics[width=0.9\textwidth]{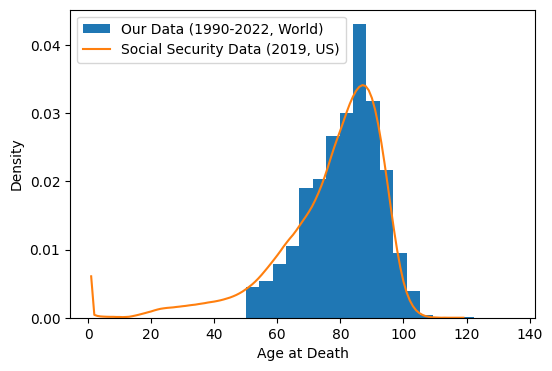}
\caption{Age at death distribution}
\label{fig:age-at-death-dist}
\end{figure}

Another source of error in our data is that we collect the images over a 33 year period (1990-2022) and from all countries. Over the years and across the countries, life expectancy changes as a result of improvements to medical technology and access to healthcare. However, as shown in \figurename~\ref{fig:age-at-death-dist}, the distribution of age at death in the data (containing 33 years and multiple countries) and the age at death in the US in 2019 (according to Social Security Administration data) are very similar. The concern about significant changes in life expectancy is the reason we did not collect data for deaths before 1990.

\begin{figure}
  \centering
  \begin{tabular}{ccc}
    \includegraphics[width=0.33\linewidth]{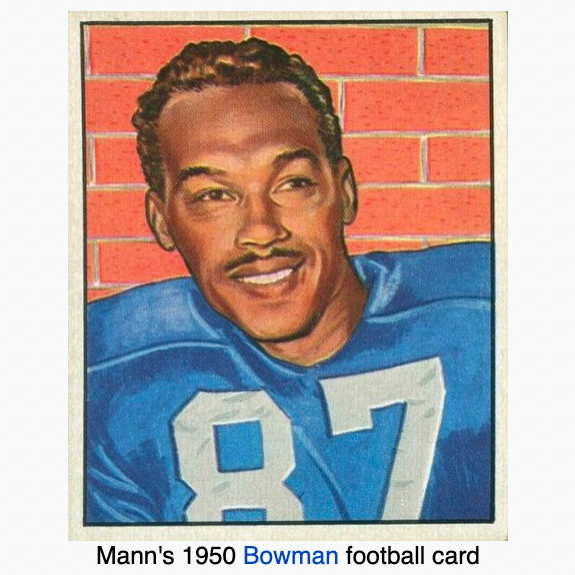} &
    \includegraphics[width=0.33\linewidth]{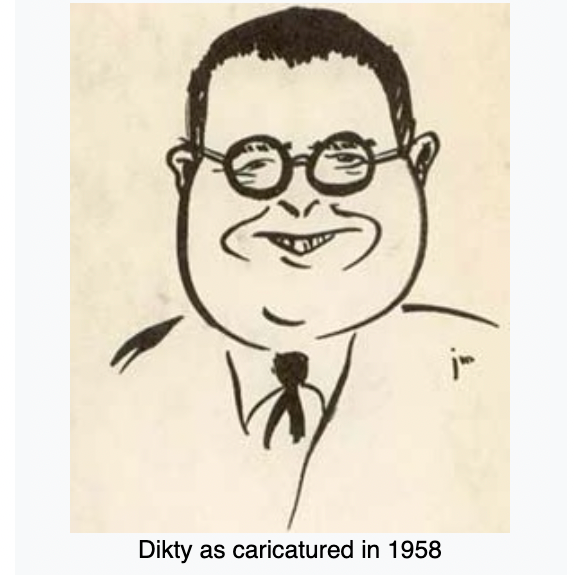} &
    \includegraphics[width=0.33\linewidth]{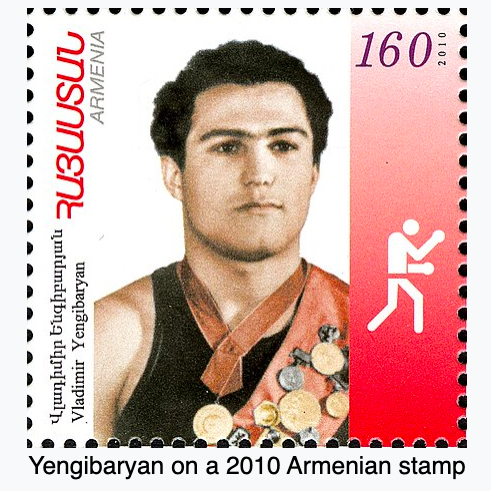}
  \end{tabular}
  \caption{Examples of Wikidata/Wikipedia entries that lead to faulty data}
  \label{fig:bad-examples}
\end{figure}

For some Wikidata/Wikipedia entries, the facial image may not be a photograph of the person. For example, the image could be a drawing or an illustration of the person's face. Furthermore, the scraped number from the image caption may not reflect the actual year the image was taken. For example, a person's Wikipedia page may include the picture of a commemorative stamp with the face of the person from a younger age, while the caption mentions the year the stamp was issued. \figurename~\ref{fig:bad-examples} displays three instances of entries that lead to erroneous data. Based on a manual inspection of 200 random entries, we found that in less than 2\% of cases the scraped images or their years were inaccurate.

After scraping Wikidata/Wikipedia, we obtained about 45,000 images and their respective labels (RL). We used Multi Task Cascaded Convolutional Neural Network (MTCNN) \citep{zhang2016joint} to detect faces in the images. We excluded those that did not contain any faces, contained more than one face, or when the face detection algorithm was less than 98\% confident that it found a human face.

After detecting the faces in the images, we aligned them by rotating the face so that the line connecting the two eyes is parallel with the horizon. We cropped the face in such a way that the distance between the eyes spans the middle 28\% of the cropped image and the eyes are 43\% from the top of the image. We also applied a simple algorithm to ensure that the pose in the image is at least partially frontal (i.e. both eyes are fully visible). While convolutional neural networks are effective at handling different poses, we wanted to make it easier for our model to extract features useful for RL prediction. To avoid distorting important information in the image, we cropped the faces in the shape of a square to match the input shape of most models. Additionally, we filtered out face crops whose width was less than 64 pixels as these images were too low quality for our model to learn from. After all the filtering, we were left with 24,167 examples.

\section{Methodology}
\label{sec:method}

To identify the most suitable model for RL prediction, we fine-tune CNNs that have been pre-trained on very large datasets for face recognition. These models map an image of a face into a high-dimensional numerical vector, with the aim that the vectors created for different images of the same person are close (as measured by cosine similarity or Euclidean distance), while those of two different people are far apart. This vector is referred to as a face embedding. We utilize these face embeddings as features in our RL prediction model.

In particular, the flattened embedding is followed by several fully-connected layers with rectified linear unit (ReLU) activation. To avoid over-fitting, both the embedding and the output of each of the fully-connected layers are passed through dropout layers. The model ultimately culminates in an output layer.

For the final (output) layer, we consider three options. The first option is a classification as defined by the argmax of a softmax layer. However, since RL is a rank variable and a classification task does not take the order of the labels into account, we believe this to be an improper way to model RL. The second option is to use the softmax probabilities ($p_i$) for labels ranging from 0 to 100 ($L_i$) and calculate the RL as the expected value ($RL = \sum_{i=0}^{100}L_i*p_i$). Lastly, the third option is to use a regression, that is, a fully connected layer of size one that generates the output. Our experiments showed that the regression works best with our data.

To select the most appropriate model to fine-tune, we evaluated FaceNet, VGGFace (a version of the VGG16 trained on a dataset of face images), and VGGFace2 (ResNet-50 trained on the VGGFace2 dataset). We conducted a simple and rapid experiment to determine which model's face embeddings were most suitable for our particular task. We froze the weights of these models and added two FC layers of 64 and 32 units, followed by an output layer of 1 unit. We trained each model for 10 epochs and compared the results.

\begin{figure}
  \centering
  \begin{tabular}{cc}
    \includegraphics[width=0.50\linewidth]{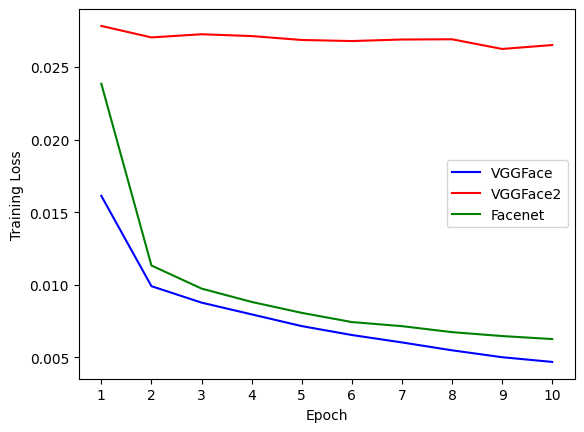} &
    \includegraphics[width=0.50\linewidth]{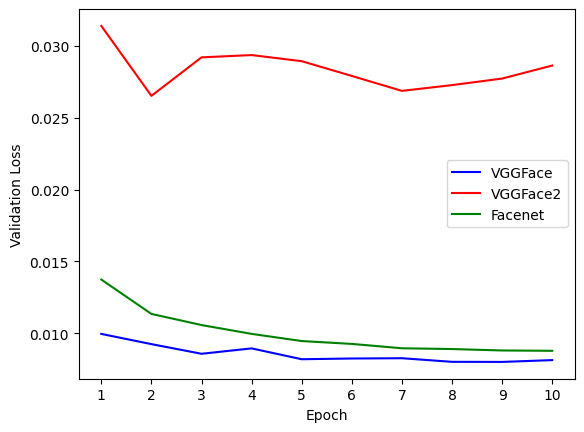} 
  \end{tabular}
  \caption{Comparing the performance of VGGFace, VGGFace2, and FaceNet embeddings for RL prediction was a preliminary step to find the model that we would use for further experimentation.}
  \label{fig:compare-models}
\end{figure}

As shown in \figurename~\ref{fig:compare-models}, VGGFace had the best performance among the three models. VGGFace (VGG16) has been successfully used for similar tasks such as age estimation in the past. VGGFace2 did not perform well, as expected, since it was specifically designed and trained to be age-invariant (i.e., to recognize a person as they age). Therefore, the embeddings generated by it do not change drastically with age, which is not beneficial for our purposes. As a result, we selected VGGFace as the basis for building our RL prediction model.

After choosing VGGFace as the embedding-generator model, we conducted experiments with different configurations. We experimented with various numbers of FC layers to be added on top of it (ranging from 1 to 3) before the output layer. We also tried different methods for generating the output (expected value, regression), different layer sizes (128, 512, 1024, and 4096), and various hyperparameters such as learning rate and batch size to optimize the performance of the model.

To fine-tune the model, we first train only the added layers while keeping the transferred weights frozen (linear probing). Then, we incrementally unfreeze and train additional layers with lower learning rates. This approach, known as "fine-tuning with freezing", allows us to leverage the pre-trained weights in the initial layers and fine-tune the model on our specific task, while avoiding the risk of catastrophic forgetting. \citet{https://doi.org/10.48550/arxiv.2202.10054} study 10 distribution shift datasets and show that linear probing followed by fine-tuning outperforms (out of distribution) just linear probing or just full fine-tuning.

\begin{figure}
\centering
\includegraphics[width=0.9\textwidth]{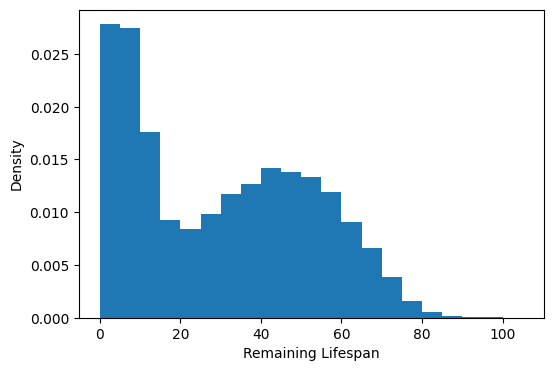}
\caption{Distribution of RL in the data}
\label{fig:rl-dist}
\end{figure}

We divide the sample into training and validation sets with a 70/30 split. As seen in \figurename~\ref{fig:rl-dist}, RL is not uniformly distributed in the data. We have relatively more examples of short RL (RL < 10), and very few examples for long RL (RL > 70). To prevent the model from becoming biased towards the majority, we divide the sample into 15 bins, with the RL being between [0,5), [5, 10), ..., [70, $\infty$). When training the model (but not for validation), we oversample the bins whose number of examples is fewer than the bin with the highest number of examples. This way, in each epoch, we will have a virtually similar number of examples for different RL levels. To clarify, these bins are only used for data balancing, not for classification. 

Before being used for training, the images are preprocessed by being scaled (to a float between 0 and 1) and resized to (244, 244). The dataset includes both colored and grayscale images. While color is likely to be relevant for RL estimation, we convert all images to grayscale to make the dataset consistent. Since the model requires a 3-channel input, the single channel of grayscale images is repeated 3 times.

\begin{figure}
  \centering
  \begin{tabular}{ccc}
    \includegraphics[width=0.33\linewidth]{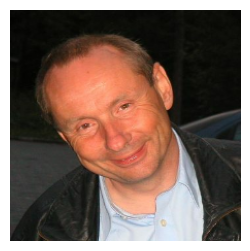} &
    \includegraphics[width=0.33\linewidth]{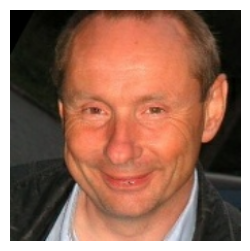} &
    \includegraphics[width=0.33\linewidth]{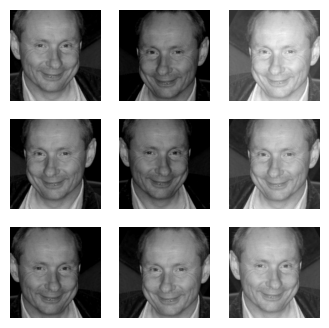} 
  \end{tabular}
  \caption{Original image, after cropping the face and alignment, and after converting to grayscale and augmentation.}
  \label{fig:augment}
\end{figure}

To augment training images, we apply random light adjustments and random horizontal flips to the images. This is followed by a random crop of (224, 224), which is the required input size for the VGGFace model. \figurename~\ref{fig:augment} illustrates an original image (fetched from Wikipedia), the image resulting from cropping the face and alignment (as described in Section \ref{sec:data}), and a sample of augmented images that are used in training. Data augmentation increases the size of the dataset (especially for the bins with fewer number of examples) and introduces variations to the images, which can help the model generalize better to unseen data.

Validation images are only converted to grayscale and resized to (224, 224). This is done to keep the validation set as similar as possible to the real-world scenarios, where the model will be applied.

The models are trained using the Huber loss function. Instead of mean Squared Error, we use Huber loss function because it is less sensitive to outliers, meaning it will not put too much weight on the difficult cases, which may be the result of faulty data. Huber loss function considers both the Mean squared error and Mean absolute error, it gives less weight to outliers and more weight to the samples that are closer to the predicted output. This helps in preventing the model from overfitting and gives a more robust and accurate prediction.

The models were trained on Google Colab GPUs using the Keras/TensorFlow framework. 

\begin{figure}
\centering
\includegraphics[width=0.9\textwidth]{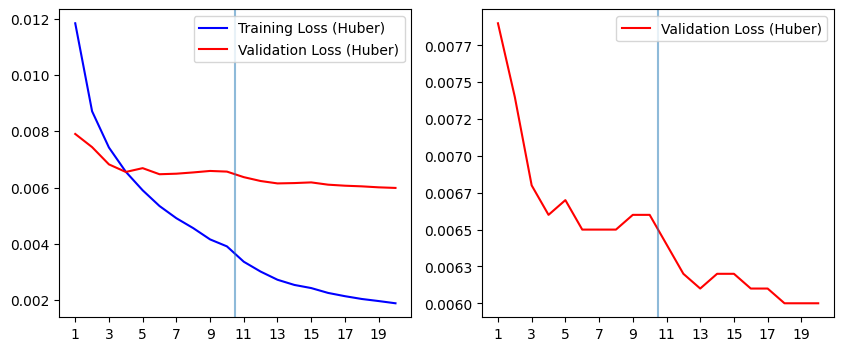}
\caption{The progression of training and validation loss over epochs. The vertical line indicates when the last 2 convolutional layers became unfrozen. }
\label{fig:epochs-loss}
\end{figure}

\begin{figure}
\centering
\includegraphics[width=0.9\textwidth]{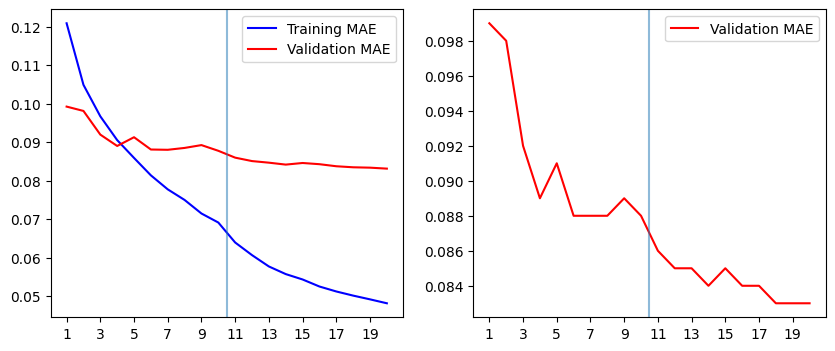}
\caption{The progression of training and validation MAE over epochs. The vertical line indicates when the last 2 convolutional layers became unfrozen. }
\label{fig:epochs-mae}
\end{figure}

After trying different architectures and hyper-parameters, the VGGFace embeddings followed by 2 FC layers of 1024 units and ending with a regression layer of 1 unit achieved the best performance (MAE of 8.3 years on validation data). In Figures \ref{fig:epochs-loss} and \ref{fig:epochs-mae}, the loss and MAE history for the training and validation sets are provided. For epochs 1 to 10, only the layers added on top of the embeddings were trained. For epochs 11 to 20 (after the vertical line), two additional convolutional layers were unfrozen and trained. However, unfreezing additional layers after that did not improve the performance. This indicates that the model has achieved a good level of accuracy and further training of the model will not improve the performance. The final model can be used to predict the RL on unseen data.

When comparing the loss and MAE between the training and validation sets, keep in mind that because of data-balancing in the training set, large RLs, which are inherently difficult cases, are oversampled. This is why the training loss/MAE starts off higher than the validation. However, as a result of oversampling, the images with large RL in the training set are augmented repetitions of a small set of originals. Thus, the model learns to perform well with them quickly, but that learning does not generalize well to the large RL cases in the validation set. This is why the training loss/MAE gets lower than those of the validation set after a few epochs.

\section{Error Analysis}
\label{sec:error-analysis}

\begin{figure}
\centering
\includegraphics[width=0.8\textwidth]{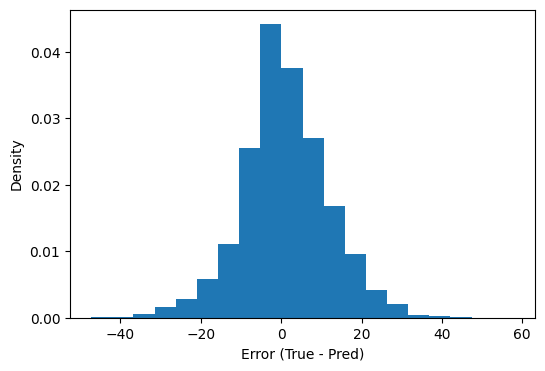}
\caption{Distribution of prediction errors (true - pred) in the validation set}
\label{fig:err-dist}
\end{figure}

To understand which cases the model has the most difficulty predicting RL, we predict the RL values for the validation set and analyze the errors. In \figurename~\ref{fig:err-dist}, the distribution of errors (true RL - predicted RL) is depicted. The figure shows that the errors are concentrated around zero, which suggests that the model is not making any systematic errors, which could be due to a problem in data or model architecture. The symmetrical distribution of errors on both sides indicates that the model is making errors of similar magnitude in both directions, over-predicting and under-predicting RL equally.

\begin{figure}
\centering
\includegraphics[width=0.8\textwidth]{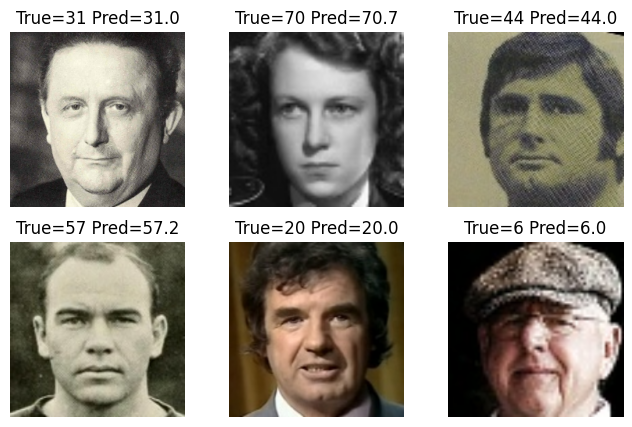}
\caption{Examples of validation images were the model's predictions are accurate}
\label{fig:good-preds}
\end{figure}

\begin{figure}
\centering
\includegraphics[width=0.8\textwidth]{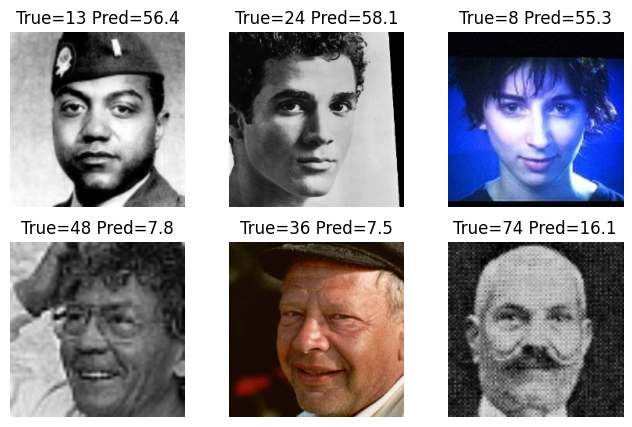}
\caption{Examples of validation images were the model's predictions are far off}
\label{fig:bad-preds}
\end{figure}

In Figures \ref{fig:good-preds} and \ref{fig:bad-preds}, we provide examples of images from the validation set where the model performs very well and images where the model performs poorly. Examination of the demonstrated "success" examples showed that they all had correct labels, and the pose in these pictures is almost completely frontal. In the demonstrated  "failure" examples, the first (top left) and last (bottom right) one turned out to have incorrect labels. In the second image of the top row, the person died of HIV/AIDS (when it was still a pandemic in early 1990s) at a young age. In the second and third images of the bottom row, the faces have a three-quarter pose. We can conclude that by improving the data through manual removal of images with erroneous labels and by increasing the number of examples with variant poses, we are likely to improve performance.

\begin{figure}
  \centering
  \begin{tabular}{cc}
    \includegraphics[width=0.50\linewidth]{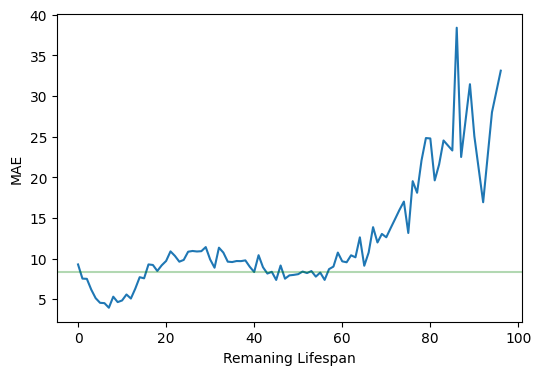} &
    \includegraphics[width=0.50\linewidth]{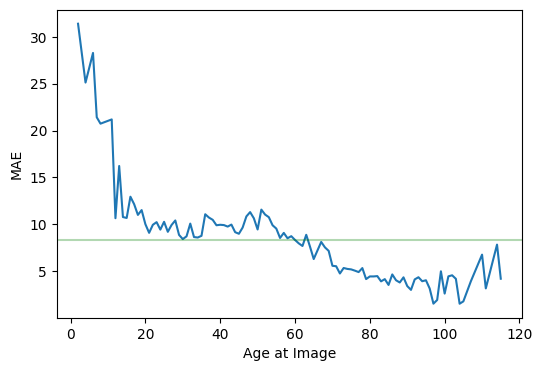} \\
    \includegraphics[width=0.50\linewidth]{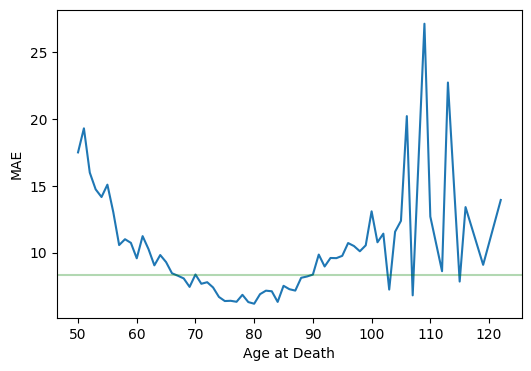} &
    \includegraphics[width=0.50\linewidth]{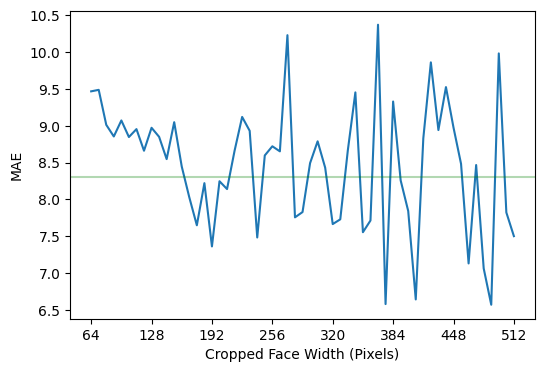}
  \end{tabular}
  \caption{MAE calculated for different levels of RL, age at image, age at death, and image width. The horizontal line represents the MAE for the entire validation set.}
  \label{fig:mae-against-vars}
\end{figure}

Figure \ref{fig:mae-against-vars} illustrates the relationship between MAE and several variables in the validation set, including:
\begin{itemize}
\item RL
\item Age at image
\item Age at death
\item Original (before resizing) image width
\end{itemize}
The horizontal line in the figure represents the MAE for the entire validation set (8.3), providing a reference point for determining whether the model's performance is better or worse than the average for specific variable values.

The MAE-RL graph illustrates that the model exhibits its best performance when the person lived less than 20 years after the image was taken. Conversely, when the actual RL is very large (the person died more than 60 years after the image was taken), the model's predictions exhibit large errors.

Similarly, the MAE vs. Age at Image graph indicates that when the person in the image is very young (less than 20 years old), the model performs poorly. This behavior is anticipated as predicting the RL of a young person poses an intrinsically difficult challenge. Moreover, the dataset contains relatively fewer images of young people, which exacerbates the problem.

The MAE vs Age at Death graph illustrates that the model's performance is suboptimal when the person in the image passed away at a very young or very old age. The model exhibits its best performance for individuals who died between the ages of 70 and 90 years old. This suggests that the model may have difficulty predicting RL for individuals outside of this age range, which could be due to a lack of training examples in those age ranges or the intrinsic difficulty of predicting RL for those age groups.

Lastly, the MAE vs. Cropped Face Width graph indicates that there is no apparent correlation between the size of the images (before being resized to match the model's required input size) and MAE. This suggests that the model's performance is not significantly affected by the width of the face crop images. However, it is important to note that this does not necessarily mean that the image quality is not important as it could be the case that the width is not a good metric of image quality.

\section{Applications}
\label{fig:apps}

\begin{figure}
\centering
\includegraphics[width=0.8\textwidth]{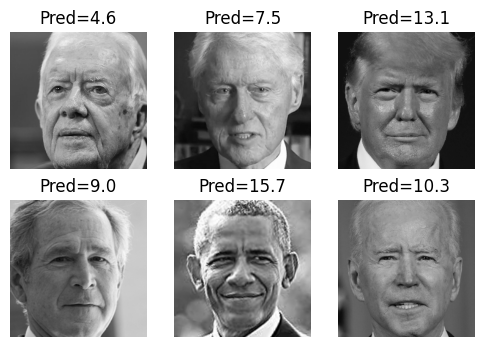}
\caption{RL predictions of the model for the living U.S. presidents using their year 2022 photographs}
\label{fig:prez}
\end{figure}

RL prediction using facial images has various potential applications. One example is that life insurance companies could use a facial RL model in conjunction with their traditional methods to obtain a more accurate understanding of a person's life expectancy and set premiums accordingly. As seen in Figure \ref{fig:prez}, the model can estimate the RL of living US presidents by using recent photographs of them.

\begin{figure}
\centering
\includegraphics[width=0.8\textwidth]{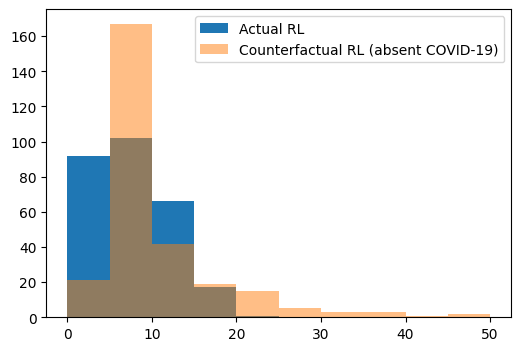}
\caption{Actual RL of a sample of individuals who died of COVID-19 vs. the RL prediction of the model had the pandemic not happened, using their photographs taken since 2000. Those individuals would have lived on average 3.3 years longer.}
\label{fig:covid}
\end{figure}

The model can also be used to estimate the loss of life (in years) for victims of fatal accidents or pandemics. As an experiment, we applied the model to predict the RL of 278 individuals collected from Wikidata/Wikipedia whose sole cause of death was recorded as COVID-19 and whose image was taken after the year 2000. It is important to note that individuals whose cause of death included COVID-19 were excluded from the training and validation datasets. Figure \ref{fig:covid} shows the distribution of the actual RL (the years between when the image was taken and when they died of COVID-19) and their predicted RL (how much they would have lived if the pandemic did not occur). It is apparent that the predicted RL distribution is to the right of the actual RL distribution. Our analysis found that, on average, these individuals would have lived 3.3 more years had the pandemic not occurred. This experiment illustrates the potential use of the model in assessing the impact of pandemics on human life expectancy.

\begin{figure}
\centering
\includegraphics[width=0.8\textwidth]{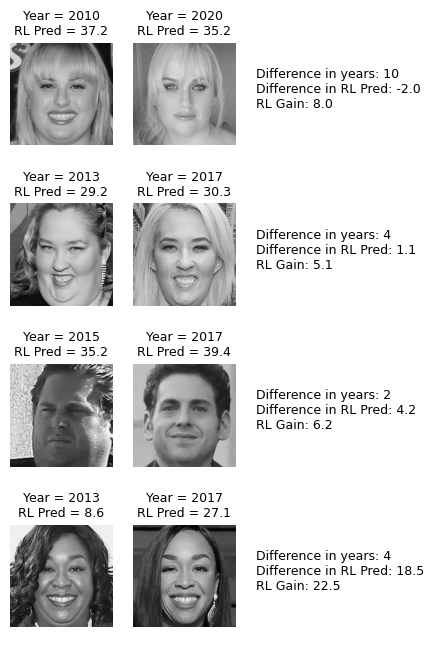}
\caption{RL gain in celebrities as a result of weight loss}
\label{fig:celeb}
\end{figure}

Another potential use of the model is to evaluate the impact of lifestyle changes and assess the effectiveness of different health interventions by measuring how much they increase the RL. To demonstrate this, we applied the model to the images of 5 celebrities before and after significant weight loss, obtained from an online article\footnote{"The craziest celebrity weight loss transformations of all time" from Page Six  \href{https://pagesix.com/article/the-craziest-celebrity-weight-loss-transformations-of-all-time/}{Link to the article}}. As seen in Figure \ref{fig:celeb}, the model suggests that these individuals increased their RL after the weight loss (considering the time between when the two images were taken).

Additionally, the model could be used for demographic studies and research in aging and longevity, providing valuable insights into the aging process.

\section{Ethical Considerations}
\label{sec:ethics}

Death is a complex and unpredictable process, and it is unlikely that we will ever be able to build models that can predict RL from the face with a very high degree of accuracy. Regardless, the ethical implications of making such predictions should be taken into consideration. There is a possibility that it could cause harm or distress to the individuals being predicted, as well as potential privacy concerns.

As mentioned previously, RL prediction models (although not using facial images) are already in use, explicitly or implicitly, in industries such as life insurance and reverse mortgage, which rely on making precise estimates about a client's RL in order to be profitable.

The argument that having a rough estimate of one's RL causes distress is similar to the case made against 23andMe's "health risk report," which provided predictions on whether a person would develop diseases such as Alzheimer's or Parkinson's based on their DNA. In that case, the FDA granted authorization for the company to provide the report to clients who opt into it, after taking into account the potential benefits and risks of providing this information to clients.

Having a rough estimate of one's own RL could make the person more cognizant of their limited time and lead them to live a more fulfilling life or change unhealthy habits in order to increase their RL. It could be used to identify individuals at high risk of mortality, enabling early interventions and improved health outcomes. 

\section{Conclusion}
\label{sec:conclusion}
In summary, predicting RL using only a person's face is a novel task that has not been attempted before, primarily due to a lack of data. To overcome this challenge, we constructed a dataset that can be used for this purpose and made it publicly available for other researchers to experiment on. By fine-tuning the VGGFace model, we achieved an MAE of approximately 8.3 years, which we believe to be satisfactory given the highly unpredictable nature of death. However, it is worth noting that the model's accuracy decreases for images of young individuals.

We also demonstrated a few potential applications of the model, including estimating the loss of life due to accidents or pandemics, or estimating the gains in RL after positive lifestyle changes. 

Overall, using AI for RL prediction solely from the face is an inherently challenging task that has the potential to benefit businesses and individuals. Further research in this area, especially using more  comprehensive and diverse data, can provide valuable insights into the aging process and help to improve the longevity of humankind.

\bibliographystyle{unsrtnat}
\bibliography{references}  %%% Uncomment this line and comment out the ``thebibliography'' section below to use the external .bib file (using bibtex) .





% \end{thebibliography}

\end{document}